\documentclass{article}

 \usepackage[final]{neurips_mlsb_2024}
\usepackage[utf8]{inputenc} 
\usepackage{caption} 
\usepackage[T1]{fontenc}    
\usepackage{hyperref}       
\usepackage{url}            
\usepackage{booktabs}       
\usepackage{amsfonts}       
\usepackage{nicefrac}       
\usepackage{microtype}      
\usepackage{xcolor}         

\usepackage{amsmath, amssymb, amsthm}
\usepackage{bm}            
\usepackage{graphicx}      
\usepackage{hyperref}      
\usepackage{enumerate}     
\usepackage{algorithm}     
\usepackage{algpseudocode} 
\usepackage{subcaption}    
\usepackage{xcolor}        
\usepackage{thmtools}      
\usepackage{cleveref}      

\declaretheorem[name=Theorem,numberwithin=section]{theorem}

\declaretheorem[name=Proposition,numberlike=theorem]{proposition}

\numberwithin{equation}{section}

\usepackage{booktabs}
\title{Relaxed Sequence Sampling for Diverse Protein Design}

\newcommand{\UMass}{\textsuperscript{1}}
\newcommand{\Flagship}{\textsuperscript{2}}
\newcommand{\intern}{\textsuperscript{*}}     
\newcommand{\senior}{\textsuperscript{\dag}}  

\author{%
\begin{tabular}{@{}c@{}}
Joohwan Ko\UMass\Flagship\intern \quad
Aristofanis Rontogiannis\Flagship \quad
Yih\mbox{-}En Andrew Ban\Flagship \\
Axel Elaldi\Flagship\senior \quad
Nicholas Franklin\Flagship\senior
\end{tabular}
}

\begin{document}

\maketitle
\begingroup
\renewcommand\thefootnote{}
\footnotetext{\footnotesize
\textsuperscript{*}\, Work was done during an internship with Flagship Pioneering.
\textsuperscript{\dag}\, Equal senior authors.
\textsuperscript{1}\,University of Massachusetts Amherst.
\textsuperscript{2}\,Flagship Pioneering.
Correspondence: \texttt{joohwanko@cs.umass.edu}.
}
\endgroup

\begin{abstract}
Protein design using structure prediction models such as AlphaFold2 has shown remarkable success, but existing approaches like relaxed sequence optimization (RSO) rely on single-path gradient descent and ignore sequence-space constraints, limiting diversity and designability. We introduce Relaxed Sequence Sampling (RSS), a Markov chain Monte Carlo (MCMC) framework that integrates structural and evolutionary information for protein design. RSS operates in continuous logit space, combining gradient-guided exploration with protein language model–informed jumps. Its energy function couples AlphaFold2-derived structural objectives with ESM2-derived sequence priors, balancing accuracy and biological plausibility. In an in silico protein binder design task, RSS produces 5$\times$ more designable structures and 2–3$\times$ greater structural diversity than RSO baselines, at equal computational cost. These results highlight RSS as a principled approach for efficiently exploring the protein design landscape.
\end{abstract}

\section{Introduction}
Protein design has transformative potential for engineering novel enzymes and targeted protein binders, yet the combinatorial sequence–structure space makes targeted design challenging \cite{kortemme2024}. To address this complexity, deep learning-based approaches fall into two complementary paradigms. The first employs large generative models to sample {\em de novo} protein sequence and/or structure \cite{watson2023novo,ingraham2023illuminating,geffner2025laproteina}. The second \emph{inverts} protein-folding models by optimizing sequences to maximize confidence for desired conformations or interfaces, recently realized through relaxed sequence optimization (RSO) and AlphaFold2 (AF2)-based binder hallucination \cite{frank2024scalable,pacesa2024bindcraft,dauparas2022robust,evans2021protein,abramson2024accurate}. While robust, this approach traces a single gradient trajectory and obtaining a diverse, high-quality candidate set is a challenge.

We introduce \emph{relaxed sequence sampling} (RSS), a Markov chain Monte Carlo (MCMC) method that preserves the robustness of inversion approaches while generating structural and sequence diversity within a \emph{single} iterative run. Our approach combines gradient-informed \emph{walks} via Metropolis-adjusted Langevin dynamics (MALA) with protein language model (PLM)-guided \emph{jumps} through a principled mixture kernel. The key component is \emph{Soft-PLM}, a differentiable relaxation enabling PLM priors to operate directly on continuous logit representations. The resulting energy couples AF2-derived structural objectives with evolutionary sequence priors, yielding substantially more diverse, high-quality candidates for the same computational budget.

\section{Preliminaries}
\paragraph{Relaxed Sequence Optimization} Relaxed sequence optimization (RSO) \cite{frank2024scalable} leverages deep learning–based protein folding methods \cite{jumper2021highly,mirdita2022colabfold,lin2023evolutionary,ahdritz2024openfold,bytedance2025protenix,abramson2024accurate} by differentiating sequences with respect to a folded conformation loss. The key idea is to replace discrete amino acid sequences with continuous logit $\ell \in \mathbb{R}^{L \times K}$, where $L$ is sequence length and $K$ is the vocabulary size. Given a target protein and desired binding interface, RSO minimizes a composite loss $\mathcal{L}_{\mathrm{AF2}}(\ell)$ combining AF2-derived metrics such as predicted local distance difference test (pLDDT), predicted aligned error (PAE), or interface contact objectives. Optimization proceeds via stochastic gradient descent on $\ell$, enabling practical protein design with frameworks like ColabDesign and providing the basis for later methods \cite{cho2025boltzdesign1,pacesa2024bindcraft}. 


\begin{figure}[t]
  \centering
  \includegraphics[width=1.0\linewidth]{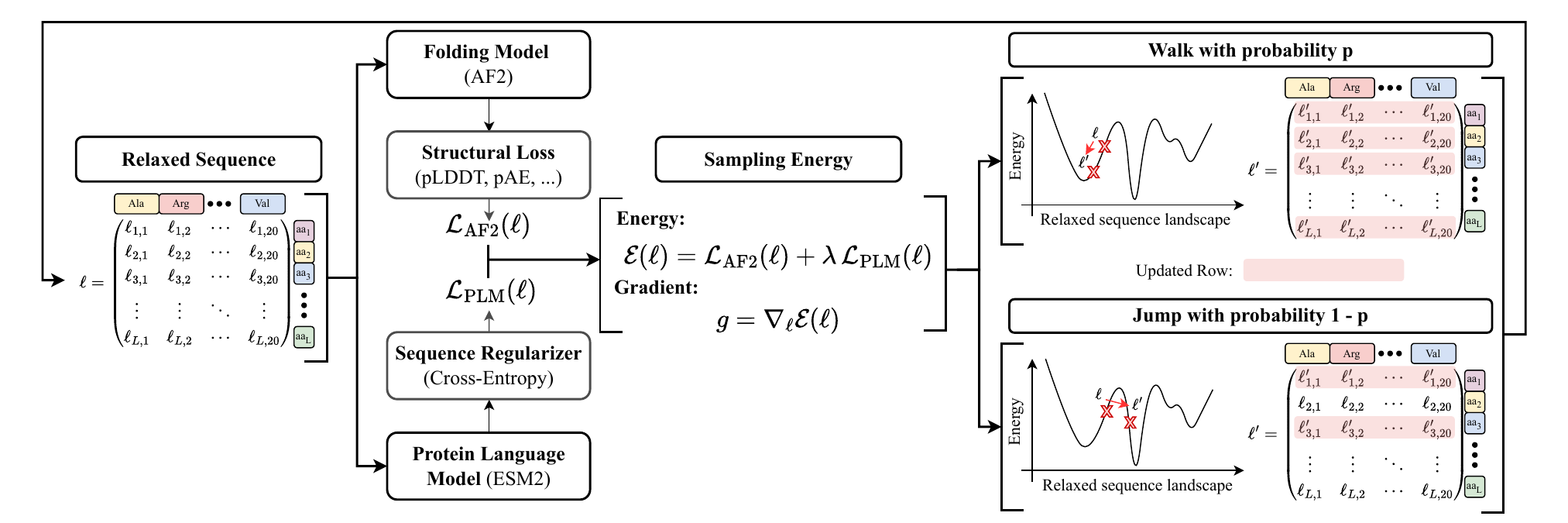}
  \caption{\textbf{Relaxed Sequence Sampling Framework.} RSS combines AF2 structural objectives with PLM priors in energy $\mathcal{E}(\ell) = \mathcal{L}_{\mathrm{AF2}}(\ell) + \lambda \mathcal{L}_{\mathrm{PLM}}(\ell)$. The walk–jump MCMC sampler alternates between gradient-informed walks and PLM-guided jumps for exploration of diverse protein designs.}
  \label{fig:rss_framework}
\end{figure}

\paragraph{Sampling for Protein Design} Sampling-based protein design has long relied on Monte Carlo methods over physics-based energies \cite{voigt2000trading,kuhlman2003design}, fragment assembly \cite{simons1997assembly}, and Monte Carlo inversion of structure prediction models \cite{anishchenko2021novo,norn2021protein}. Recent efforts combine gradient descent with MCMC, assuming that stochasticity aids escape from local minima. BAGEL frames design as sampling over a modular energy landscape using Monte Carlo optimizers and ML oracles, operating in discrete amino acid sequence space \citep{lala2025bagel}. Concurrently, protein language models (PLMs) like ESM-2 trained on amino acid sequences, provide powerful evolutionary priors \cite{rives2021biological,lin2023evolutionary,rao2021msa,meier2021language}, complementing structure-based objectives, and showing promise within structure-prediction models \cite{lin2023evolutionary, wang2025simplefoldfoldingproteinssimpler, lu2025allatom, hayes2025simulating}. 

Nonetheless, current methods face a fundamental trade-off: they either demand the computational resources for multiple trajectories or struggle to effectively balance exploration and exploitation. Our approach addresses this limitation by unifying efficient gradient-based moves with PLM-guided jumps within continuous logit space, enabling both principled exploration and computational efficiency.

\section{Method}
We present \emph{Relaxed Sequence Sampling} (RSS), a novel MCMC method operating over continuous amino acid logit to explore a structure- and sequence-aware energy landscape. To overcome single-trajectory optimization limitation, RSS employs a walk–jump sampling strategy balancing local gradient-based exploration with global PLM-guided moves. RSS key insight is combining two information: (1) an AF2-derived structural objective computed directly from continuous logits, guiding the sampler toward sequences with desired  predicted conformations, and (2) a differentiable PLM prior constraining exploration to evolutionarily plausible sequence regions. This dual-objective design enables efficient sampling across energy barriers while maintaining biological realism.

\paragraph{Notation.}
We work with logit matrices $\ell\in\mathbb{R}^{L\times K}$ where $L$ is sequence length and $K$ is vocabulary size. Each row $\ell_i\in\mathbb{R}^K$ yields relaxed per-site marginals $q_i=\mathrm{softmax}(\ell_i)\in\Delta^{K-1}$. We use $W^{\mathrm{AA}}\in\mathbb{R}^{K\times d}$ for the amino-acid embedding table and $w_{\mathrm{mask}}\in\mathbb{R}^d$ for the PLM mask embedding.

\subsection{Energy Formulation and Target Distribution}

We formulate protein design as sampling from a Boltzmann distribution over continuous logit space $\pi(\ell)\propto\exp\!\big(\text{-}\beta\mathcal{E}(\ell)\big)$. We define a composite energy  $\mathcal{E}(\ell)=\mathcal{L}_{\mathrm{AF2}}(\ell)+\lambda\,\mathcal{L}_{\mathrm{PLM}}(\ell)$ that balances structural objectives with evolutionary sequence priors, with weights $\lambda,\beta>0$. Rather than following a single optimization trajectory, our sampler generates diverse relaxed sequences by exploring low-energy regions. The energy combines an AF2-derived structural term $\mathcal{L}_{\mathrm{AF2}}(\ell)$ guiding toward accurate folded sequences with a PLM-based term $\mathcal{L}_{\mathrm{PLM}}(\ell)$ ensuring evolutionary plausibility.

\paragraph{AF2-derived structural objective.}
The structural component leverages AF2's differentiable architecture, directly processing continuous logit representations $\ell$. We construct a composite structural loss $\mathcal{L}_{\mathrm{AF2}}(\ell)$ incorporating multiple AF2-derived metrics including pLDDT for structural confidence, PAE for inter-residue accuracy, interface contact objectives for binding specificity, and optional geometric restraints. Gradients $\nabla_\ell \mathcal{L}_{\mathrm{AF2}}(\ell)$ are computed via backpropagation through the full AF2 forward pass executed on the continuous logits $\ell$. This integration embeds the full structural model in the sampling loop, enabling proposals guided by the local curvature of the AF2 objective landscape.

\begin{algorithm}[t]
\caption{Relaxed Sequence Sampling (RSS)}
\label{alg:rss}
\begin{algorithmic}[1]
\State \textbf{Input:} energy $\mathcal{E}(\ell)=\mathcal{L}_{\mathrm{AF2}}(\ell)+\lambda\,\mathcal{L}_{\mathrm{PLM}}(\ell)$; inverse temperature $\beta$; PLM weight $\lambda$; MALA step $\eta$; jump probability $p_{\mathrm{jump}}$; mask scale $\kappa$; swap size $\gamma$; PLM temperature $\tau$; number of iterations $T$
\State Initialize logits $\ell_0$; compute $E_0=\mathcal{E}(\ell_0)$ and $g_0=\nabla_{\ell}\mathcal{E}(\ell_0)$
\For{$t=0,\ldots,T-1$}
  \State Draw $u\sim \mathrm{Uniform}(0,1)$
  \If{$u>p_{\mathrm{jump}}$} \Comment{Walk (MALA)}
    \State Propose $\ell'=\ell_t-\eta\,g_t+\sqrt{2\eta/\beta}\,\xi$, with $\xi\sim\mathcal{N}(0,I)$
    \State Compute $E'=\mathcal{E}(\ell')$ and $g'=\nabla_{\ell}\mathcal{E}(\ell')$; set $\alpha=\alpha_{\mathrm{walk}}(\ell_t,\ell')$ \hfill\eqref{eq:walk-acc}
  \Else \Comment{Jump (masked-PLM swap)}
    \State Sample a nonempty $S\subseteq\{1,\ldots,L\}$ with per-site probabilities $p_i(\ell_t)$ (resample if $S=\varnothing$)
    \State For each $i\in S$: sample $y_i^{+}\!\sim p^{\mathrm{PLM}}_i(\cdot\mid \ell_t;\tau)$ and $y_i^{-}\!\sim r_i$; set $\ell'_i=\ell_{t,i}+\gamma\,(e_{y_i^{+}}-e_{y_i^{-}})$; for $i\notin S$, set $\ell'_i=\ell_{t,i}$
    \State Compute $E'=\mathcal{E}(\ell')$, $g'=\nabla_{\ell}\mathcal{E}(\ell')$; form $q_S(S\mid\ell_t)$ and $q_S(S\mid\ell')$; set $\alpha=\alpha_{\mathrm{jump}}(\ell_t,\ell')$ \hfill\eqref{eq:jump-acc}
  \EndIf
  \State Accept with probability $\alpha$: set $\ell_{t+1}\leftarrow\ell'$ if accepted; else $\ell_{t+1}\leftarrow\ell_t$
  \State Update $E_{t+1}=\mathcal{E}(\ell_{t+1})$, $g_{t+1}=\nabla_{\ell}\mathcal{E}(\ell_{t+1})$
\EndFor
\State \textbf{Output:} a collection of relaxed sequences $\{\ell_t\}$ concentrated in low-energy regions
\end{algorithmic}
\end{algorithm}

\paragraph{Soft-PLM: differentiable sequence prior.}
Our key component is Soft-PLM, enabling protein language models to operate directly on continuous logit representations without discretization. This preserves long-range sequence context while maintaining differentiability essential for gradient-based sampling.  We begin by computing 
\begin{equation}
    z_i(\ell)=\sum_{k=1}^{K} q_i[k]\;W^{\mathrm{AA}}_{k},
\end{equation} 
the expected amino acid embeddings from the relaxed marginal distributions $q_i=\mathrm{softmax}(\ell_i)$, where 
$W^{\mathrm{AA}}\in\mathbb{R}^{K\times d}$ represents the frozen amino acid embedding matrix from the PLM. For each position $i$, we construct a masked context, by replacing the $i$-th embedding with the PLM's mask token:
\begin{equation}
\tilde z^{(i)}(\ell)=[z_1,\ldots,z_{i-1},\,w_{\mathrm{mask}},\,z_{i+1},\ldots,z_L].
\end{equation}
Processing this masked context through the frozen PLM yields position-specific conditional distributions:
\begin{equation}
p^{\mathrm{PLM}}_i(\cdot\mid \ell;\tau)=\mathrm{softmax}\Big(\mathrm{logits}\big(\tilde z^{(i)}(\ell)\big)/\tau\Big)_i \in \Delta^{K-1},  \label{eq:plm-cond}
\end{equation}
where $\tau>0$ is a temperature parameter controlling the sharpness of the PLM predictions. The Soft-PLM loss encourages alignment between the relaxed marginals and PLM conditionals via cross-entropy minimization:
\begin{equation}
\mathcal{L}_{\mathrm{PLM}}(\ell)=\sum_{i=1}^{L} H\big(q_i,\ p^{\mathrm{PLM}}_i(\cdot\mid \ell;\tau)\big)
=\,-\sum_{i=1}^{L}\ \mathbb{E}_{x\sim q_i}\!\left[\log p^{\mathrm{PLM}}_i(x\mid \ell;\tau)\right]. \label{eq:soft-plm}
\end{equation}
This formulation ensures continuous sequence representations align with evolutionary patterns captured by the PLM, constraining the sampler to biologically plausible sequence space regions.

\subsection{Walk–Jump MCMC Sampling Strategy}
\begin{figure}[t]
    \centering
    \includegraphics[width=1.0\linewidth]{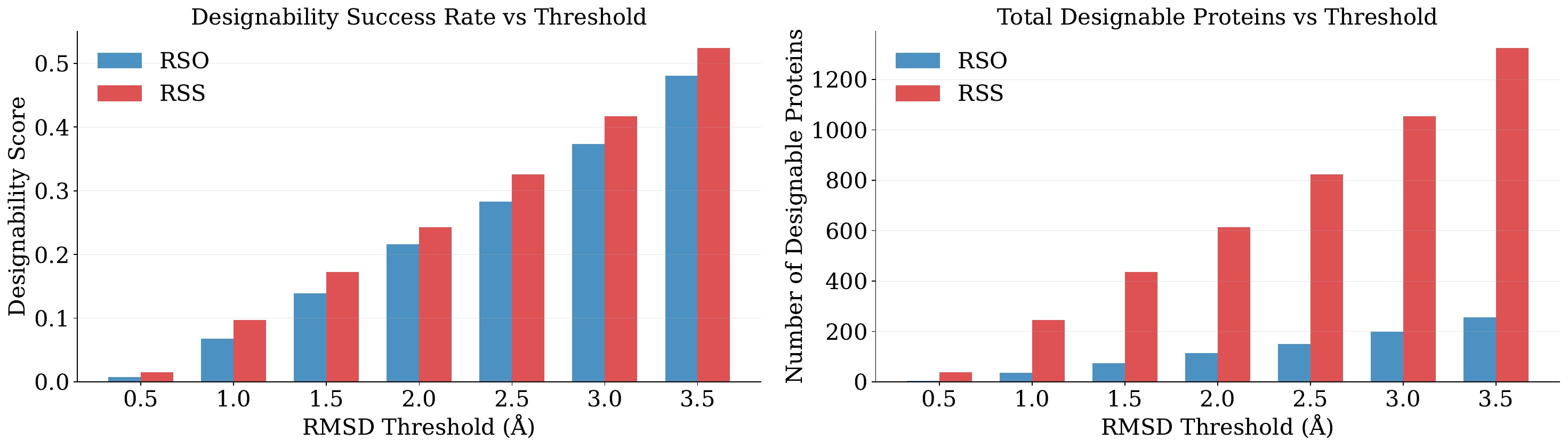}
    \caption{\textbf{RSS maintains superior performance across RMSD thresholds.} Left: Designability success rate (fraction of designable samples) as a function of RMSD threshold. Right: Total number of designable proteins as a function of RMSD threshold. RSS (red) consistently outperforms RSO (blue) across all thresholds.}
    \label{fig:designability}
\end{figure}

The core RSS idea is a $\pi$-reversible \emph{mixture transition kernel}
$\mathcal{T}=(1-p_{\mathrm{jump}})\,\mathcal{T}_{\mathrm{walk}}+p_{\mathrm{jump}}\,\mathcal{T}_{\mathrm{jump}}$,
with jump probability $p_{\mathrm{jump}}\in(0,1)$. The sampler alternates between
gradient-informed walks that exploit local curvature of $\mathcal{E}(\ell)$ and
PLM-guided jumps that propose coherent multi-site changes to cross energy barriers.
The parameter $p_{\mathrm{jump}}$ balances local refinement and global exploration.

\paragraph{Gradient-informed walks via MALA.}
The walk component employs Metropolis-Adjusted Langevin Algorithm (MALA) to leverage gradient information for efficient local exploration. Given the current state $\ell$ with energy gradient $g=\nabla_{\ell}\mathcal{E}(\ell)$ and step size $\eta>0$, we propose $\ell'=\ell-\eta\,g+\sqrt{\tfrac{2\eta}{\beta}}\xi$ with noise $\xi\sim\mathcal{N}(0,I)$. This proposal combines a deterministic drift toward lower energy regions with controlled stochastic perturbation. The acceptance probability follows the standard MALA formula:
\begin{equation}
\alpha_{\mathrm{walk}}(\ell,\ell')=\min\!\left\{1,\ \frac{\pi(\ell')\,q_{\mathrm{walk}}(\ell\mid\ell')}{\pi(\ell)\,q_{\mathrm{walk}}(\ell'\mid\ell)}\right\},
\quad q_{\mathrm{walk}}=\mathcal{N}\!\big(\ell-\eta\,\nabla \mathcal{E}(\ell),\ (2\eta/\beta)I\big). \label{eq:walk-acc}
\end{equation}
\paragraph{PLM-guided jumps for global exploration.}
The jump mechanism targets the inability of purely gradient-based updates to traverse energy barriers between distant sequence modes. Let $\ell\in\mathbb{R}^{L\times K}$ be the current logit matrix, $\mathcal{E}(\ell)$ the energy defining the target, and $g=\nabla_{\ell}\mathcal{E}(\ell)$ its gradient, with row $g_i\in\mathbb{R}^{K}$ at position $i$; $\|\cdot\|_2$ denotes the Euclidean norm. We first sample a nonempty mask $S\subseteq\{1,\ldots,L\}$ by independent Bernoulli draws with per-site probabilities
\begin{equation}
p_i(\ell)=\min\!\left\{1,\ \kappa\,\frac{\|g_i\|_2}{\max_{j}\|g_j\|_2+\varepsilon}\right\},(\kappa\in(0,1],\ \varepsilon>0),
\end{equation}
resampling until $S\neq\varnothing$. The resulting mask proposal mass is
\begin{equation}
q_S(S\mid \ell)=\prod_{i=1}^{L} p_i(\ell)^{\mathbb{I}\{i\in S\}}\big(1-p_i(\ell)\big)^{\mathbb{I}\{i\notin S\}},
\end{equation}
where $\mathbb{I}\{\cdot\}$ is the indicator function. For each $i\in S$, draw a \emph{forward} token $y_i^{+}\sim p^{\mathrm{PLM}}_i(\cdot\mid \ell;\tau)$ from the (Soft-)PLM masked conditional at site $i$ with temperature $\tau>0$, and a \emph{reference} token $y_i^{-}\sim r_i(\cdot)$ from a base distribution $r_i$ taken uniform over the vocabulary $\{1,\ldots,K\}$, i.e., $r_i(y)=1/K$. Apply the “swap” update of magnitude $\gamma>0$:
\begin{equation}
\ell'_i=\begin{cases}
\ell_i+\gamma\,(e_{y_i^{+}}-e_{y_i^{-}}), & i\in S,\\[2pt]
\ell_i, & i\notin S,
\end{cases}
\end{equation}
where $e_y\in\mathbb{R}^{K}$ is the one-hot (standard basis) vector for token $y$. Let $g'=\nabla_{\ell}\mathcal{E}(\ell')$ and define $p_i(\ell')$ and $q_S(S\mid \ell')$ analogously at the proposed state. With inverse temperature $\beta>0$ and target distribution $\pi(\ell)\propto \exp(-\beta \mathcal{E}(\ell))$, the Metropolis–Hastings acceptance probability is
\begin{equation}
\alpha_{\mathrm{jump}}(\ell,\ell')=\min\!\left\{1,\
\exp\!\big(-\beta[\mathcal{E}(\ell')-\mathcal{E}(\ell)]\big)\,
\frac{q_S(S\mid \ell')}{q_S(S\mid \ell)}\,
\prod_{i\in S}\frac{p^{\mathrm{PLM}}_i(y_i^{-}\mid \ell';\tau)}{p^{\mathrm{PLM}}_i(y_i^{+}\mid \ell;\tau)}
\right\}. \label{eq:jump-acc}
\end{equation}
The factors correspond to (left to right) the target ratio, the mask-selection proposal ratio, and the reverse/forward PLM conditionals; the uniform $r_i$ cancels in the proposal ratio, and the deterministic affine swap has unit Jacobian, ensuring detailed balance.

\paragraph{Algorithm and practicalities}
Algorithm~\ref{alg:rss} summarizes RSS. In practice we tune $\eta$ to keep MALA acceptance in a healthy range (e.g., $40$–$60\%$) and choose $\gamma$ so jumps neither collapse to no-ops nor suffer vanishing acceptance. To manage PLM/AF2 cost, we restrict $|S|$ (typically $1$–$3$ sites per jump) and cache conditionals where possible. All quantities needed in \eqref{eq:walk-acc} and \eqref{eq:jump-acc} are evaluated at both $\ell$ and $\ell'$ (including the updated mask mass $q_S(S\mid\ell')$), which is necessary for exact detailed balance.

\section{Experiments}
\begin{table}[t]
\centering
\caption{\textbf{RSS generates more designable binders than RSO.} Number of unique designable binder structures per target protein, evaluated with MPNN-1 and MPNN-8 (1 and 8 sequences from ProteinMPNN). Each method was run for 50 seeds with equal computational budget. RSO$^*$ denotes RSO without PLM guidance.}
\label{tab:designable_structures}
\resizebox{\textwidth}{!}{%
\begin{tabular}{lrrrrrr}
\toprule
Target & RSO$^*$(MPNN-1) & RSO(MPNN-1) & RSS(MPNN-1) & RSO$^*$(MPNN-8) & RSO(MPNN-8) & RSS(MPNN-8) \\
\midrule
\textbf{Average} & 13.7 & 14.4 & 76.0 & 22.7 & 25.6 & \textbf{129.0} \\
\midrule
8aed & 17 & 18 & 108 & 34 & 37 & \textbf{175} \\
8czf & 21 & 22 & 112 & 29 & 38 & \textbf{174} \\
5uuk & 13 & 14 & 81 & 25 & 25 & \textbf{147} \\
8czh & 17 & 18 & 90 & 26 & 28 & \textbf{136} \\
7z7c & 10 & 9 & 61 & 24 & 24 & \textbf{143} \\
5tzp & 14 & 19 & 96 & 25 & 32 & \textbf{151} \\
5uu1 & 22 & 21 & 85 & 23 & 26 & \textbf{114} \\
4lx3 & 3 & 0 & 13 & 9 & 5 & \textbf{50} \\
4ue8 & 11 & 12 & 53 & 22 & 22 & \textbf{98} \\
3vz9 & 9 & 11 & 61 & 10 & 19 & \textbf{102} \\
\midrule
\textbf{Total} & 137 & 144 & 760 & 227 & 256 & \textbf{1290} \\
\bottomrule
\end{tabular}%
}
\end{table}

We demonstrate that RSS yields greater diversity than RSO under equal compute budgets, then validate that Soft-PLM provides a faithful differentiable surrogate for discrete PLM operations.

\subsection{Binder Design Results}

\textbf{Setup.} We evaluate RSS on protein binder design using 10 protein complexes from the PINDER dataset \cite{pinder2024}. For each complex, we designate one chain as the target (e.g., chain A in 8aed) and design a binder with the same length as the original partner chain (e.g., chain C in 8aed). All methods optimize the same AF2-based energy combining interface pLDDT, PAE, and contact objectives.

\textbf{Baselines.} We compare RSS against Relaxed Sequence Optimization (RSO) with and without PLM guidance. To ensure fair comparison, all methods use identical computational resources: each method is run for 50 different random seeds with the same amount of GPU hours per method.

\textbf{Evaluation Metrics.} Following \cite{geffner2025laproteina}, we evaluate designability and diversity of designs. For designability, we use ProteinMPNN to generate diverse sequences for each structure and fold them with Protenix \citep{bytedance2025protenix}. A structure is designable if at least one sequence achieves RMSD < 3.5Å to the original structure. We test two variants: MPNN-1 (using 1 sequence) and MPNN-8 (using 8 sequences). For diversity, we cluster backbone coordinates using Foldseek with a 0.5 TM-score threshold.

Table~\ref{tab:designable_structures} shows that RSS dramatically outperforms RSO in generating diverse, high-quality designs. Using the MPNN-1 metric, RSS produces 5.3× more designable structures than RSO. With the more stringent MPNN-8 metric, RSS still achieves a 5× improvement, demonstrating that the gains are robust across evaluation criteria. RSO with PLM guidance (RSO) shows a marginal improvement over vanilla RSO (RSO$^*$), and RSS's walk-jump mechanism leverages the PLM effectively.

\begin{table}[t]
\centering
\caption{\textbf{RSS discovers more diverse binder clusters than RSO.} Number of distinct structural clusters among designable binder structures per target protein. Each method was run for 50 seeds with equal computational budget. RSO$^*$ denotes RSO without PLM guidance.}
\label{tab:designable_clusters}
\resizebox{\textwidth}{!}{%
\begin{tabular}{lrrrrrr}
\toprule
Target & RSO$^*$(MPNN-1) & RSO(MPNN-1) & RSS(MPNN-1) & RSO$^*$(MPNN-8) & RSO(MPNN-8) & RSS(MPNN-8) \\
\midrule
\textbf{Average} & 12.1 & 12.3 & 35.1 & 21.1 & 22.6 & \textbf{53.2} \\
\midrule
8aed & 17 & 18 & 65 & 33 & 37 & \textbf{92} \\
8czf & 17 & 17 & 61 & 27 & 32 & \textbf{86} \\
5uuk & 13 & 14 & 42 & 25 & 25 & \textbf{69} \\
8czh & 17 & 15 & 41 & 25 & 25 & \textbf{60} \\
7z7c & 9 & 7 & 34 & 22 & 22 & \textbf{63} \\
5tzp & 12 & 19 & 39 & 22 & 31 & \textbf{61} \\
5uu1 & 21 & 18 & 37 & 23 & 23 & \textbf{44} \\
4lx3 & 2 & 0 & 5 & 9 & 4 & \textbf{13} \\
4ue8 & 8 & 8 & 14 & 15 & 16 & \textbf{22} \\
3vz9 & 5 & 7 & 13 & 10 & 11 & \textbf{22} \\
\midrule
\textbf{Total} & 121 & 123 & 351 & 211 & 226 & \textbf{532} \\
\bottomrule
\end{tabular}%
}
\end{table}

Table~\ref{tab:designable_clusters} reveals the diversity of discovered structures. RSS yields 53.2 distinct clusters on average (MPNN-8 metric) versus 22.6 for RSO—a 2.4× improvement. This indicates that RSS's additional designs span genuinely different structural modes rather than minor variations of the same fold. The walk-jump mechanism enables RSS to traverse distant basins in the energy landscape that gradient descent alone cannot reach.

Figure~\ref{fig:designability} analyzes designability across different RMSD thresholds. RSS maintains its advantage across the full range of RMSD cutoffs, demonstrating that the improvements are not artifacts of a particular threshold choice. The consistent gap between RSS and RSO variants highlights the effectiveness of combining gradient-informed walks with PLM-guided jumps.

\subsection{Ablation: Soft-PLM Validation}

A key component of RSS is Soft-PLM, which enables differentiable PLM operations in continuous logit space. We validate that Soft-PLM accurately approximates a pretrained discrete protein language model (ESM2\_t30\_150M) across 500 natural proteins from UniProt.

We evaluate three aspects: (1) \textit{One-hot fidelity}: For exact one-hot contexts, Soft-PLM's masked conditionals closely match the discrete PLM (KL=0.034) with near-perfect gradient correlation (0.987). (2) \textit{Mixture consistency}: With 30\% of positions blurred, Soft-PLM accurately approximates Monte Carlo marginalization (JS=0.017, top-1 agreement=0.839). (3) \textit{Library ranking}: For mutational libraries, Soft-PLM scores correlate perfectly with discrete PLM pseudo-likelihoods (correlation=1.0).

These results (Table~\ref{tab:softesm-main}) confirm that Soft-PLM provides a faithful differentiable surrogate for discrete PLM operations, enabling effective PLM-guided jumps within RSS while maintaining the advantages of continuous optimization.

\section{Discussion}

Relaxed Sequence Sampling (RSS) replaces single-trajectory RSO with a walk–jump sampler in continuous logit space, mixing gradient-informed MALA steps with PLM-guided edits to explore more of the design landscape at matched compute. Empirically, RSS yields ~5× more designable binder backbones and ~2–3× more structural clusters (Tables~\ref{tab:designable_structures}–\ref{tab:designable_clusters}), with gains stable across RMSD thresholds (Fig.~\ref{fig:designability}), indicating genuine mode discovery rather than near-duplicates. Key to this effect are Soft-PLM, which exposes masked PLM conditionals directly on relaxed logits, and gradient-norm masking, which targets jumps to high-curvature sites. Limitations include reliance on learned surrogates (AF2 for design; ProteinMPNN/Protenix for scoring), possible over-regularization from large PLM weight $\lambda$, and evaluation on fixed-length binders. Future extensions—interleaving structure-space moves, adaptive mixing/tempering across $\beta$, and function- or MSA-conditioned priors—could further improve efficiency and practical utility.

\clearpage
\bibliographystyle{abbrv}
\bibliography{references}

\begin{thebibliography}{10}

\bibitem{abramson2024accurate}
J.~Abramson, J.~Adler, J.~Dunger, R.~Evans, T.~Green, A.~Pritzel, O.~Ronneberger, L.~Willmore, A.~J. Ballard, J.~Bambrick, et~al.
\newblock Accurate structure prediction of biomolecular interactions with alphafold 3.
\newblock {\em Nature}, 630(8016):493--500, 2024.

\bibitem{ahdritz2024openfold}
G.~Ahdritz, N.~Bouatta, C.~Floristean, S.~Kadyan, Q.~Xia, W.~Gerecke, T.~J. O’Donnell, D.~Berenberg, I.~Fisk, N.~Zanichelli, et~al.
\newblock Openfold: Retraining alphafold2 yields new insights into its learning mechanisms and capacity for generalization.
\newblock {\em Nature methods}, 21(8):1514--1524, 2024.

\bibitem{anishchenko2021novo}
I.~Anishchenko, S.~J. Pellock, T.~M. Chidyausiku, et~al.
\newblock De novo protein design by deep network hallucination.
\newblock {\em Nature}, 600(7889):547--552, 2021.

\bibitem{cho2025boltzdesign1}
Y.~Cho, M.~Pacesa, Z.~Zhang, B.~E. Correia, and S.~Ovchinnikov.
\newblock Boltzdesign1: Inverting all-atom structure prediction model for generalized biomolecular binder design.
\newblock {\em bioRxiv}, pages 2025--04, 2025.

\bibitem{dauparas2022robust}
J.~Dauparas, I.~Anishchenko, N.~Bennett, H.~Bai, R.~J. Ragotte, L.~F. Milles, B.~I. Wicky, A.~Courbet, R.~J. de~Haas, N.~Bethel, et~al.
\newblock Robust deep learning--based protein sequence design using proteinmpnn.
\newblock {\em Science}, 378(6615):49--56, 2022.

\bibitem{evans2021protein}
R.~Evans, M.~O’Neill, A.~Pritzel, N.~Antropova, A.~Senior, T.~Green, A.~{\v{Z}}{\'\i}dek, R.~Bates, S.~Blackwell, J.~Yim, et~al.
\newblock Protein complex prediction with alphafold-multimer.
\newblock {\em biorxiv}, pages 2021--10, 2021.

\bibitem{frank2024scalable}
C.~Frank, A.~Khoshouei, L.~Fu$\beta$, D.~Schiwietz, D.~Putz, L.~Weber, Z.~Zhao, M.~Hattori, S.~Feng, Y.~de~Stigter, et~al.
\newblock Scalable protein design using optimization in a relaxed sequence space.
\newblock {\em Science}, 386(6720):439--445, 2024.

\bibitem{geffner2025laproteina}
T.~Geffner, K.~Didi, Z.~Cao, D.~Reidenbach, Z.~Zhang, C.~Dallago, E.~Kucukbenli, K.~Kreis, and A.~Vahdat.
\newblock La-proteina: Atomistic protein generation via partially latent flow matching.
\newblock {\em arXiv preprint arXiv:2507.09466}, 2025.

\bibitem{hayes2025simulating}
T.~Hayes, R.~Rao, H.~Akin, N.~J. Sofroniew, D.~Oktay, Z.~Lin, R.~Verkuil, V.~Q. Tran, J.~Deaton, M.~Wiggert, et~al.
\newblock Simulating 500 million years of evolution with a language model.
\newblock {\em Science}, 387(6736):850--858, 2025.

\bibitem{ingraham2023illuminating}
J.~B. Ingraham, M.~Baranov, Z.~Costello, K.~W. Barber, W.~Wang, A.~Ismail, V.~Frappier, D.~M. Lord, C.~Ng-Thow-Hing, E.~R. Van~Vlack, et~al.
\newblock Illuminating protein space with a programmable generative model.
\newblock {\em Nature}, 623(7989):1070--1078, 2023.

\bibitem{jumper2021highly}
J.~Jumper, R.~Evans, A.~Pritzel, T.~Green, M.~Figurnov, O.~Ronneberger, K.~Tunyasuvunakool, R.~Bates, A.~{\v{Z}}{\'\i}dek, A.~Potapenko, et~al.
\newblock Highly accurate protein structure prediction with alphafold.
\newblock {\em nature}, 596(7873):583--589, 2021.

\bibitem{kortemme2024}
T.~Kortemme.
\newblock De novo protein design—from new structures to programmable functions.
\newblock {\em Cell}, 187(3):526--544, 2024.

\bibitem{pinder2024}
D.~Kovtun, M.~Akdel, A.~Goncearenco, G.~Zhou, G.~Holt, D.~Baugher, D.~Lin, Y.~Adeshina, T.~Castiglione, X.~Wang, C.~Marquet, M.~McPartlon, T.~Geffner, G.~Corso, H.~St{\"a}rk, Z.~Carpenter, E.~Kucukbenli, M.~Bronstein, and L.~Naef.
\newblock Pinder: The protein interaction dataset and evaluation resource.
\newblock {\em bioRxiv}, 2024.

\bibitem{kuhlman2003design}
B.~Kuhlman, G.~Dantas, G.~C. Ireton, et~al.
\newblock Design of a novel globular protein fold with atomic-level accuracy.
\newblock {\em Science}, 302(5649):1364--1368, 2003.

\bibitem{lala2025bagel}
J.~L{\'a}la, A.~Al-Saffar, and S.~Angioletti-Uberti.
\newblock Bagel: Protein engineering via exploration of an energy landscape.
\newblock {\em bioRxiv}, pages 2025--07, 2025.

\bibitem{lin2023evolutionary}
Z.~Lin, H.~Akin, R.~Rao, B.~Hie, Z.~Zhu, W.~Lu, N.~Smetanin, R.~Verkuil, O.~Kabeli, Y.~Shmueli, et~al.
\newblock Evolutionary-scale prediction of atomic-level protein structure with a language model.
\newblock {\em Science}, 379(6637):1123--1130, 2023.

\bibitem{lu2025allatom}
A.~X. Lu, W.~Yan, S.~A. Robinson, S.~Kelow, K.~K. Yang, V.~Gligorijevic, K.~Cho, R.~Bonneau, P.~Abbeel, and N.~C. Frey.
\newblock All-atom protein generation with latent diffusion.
\newblock In {\em ICLR 2025 Workshop on Generative and Experimental Perspectives for Biomolecular Design}, 2025.

\bibitem{meier2021language}
J.~Meier, R.~Rao, R.~Verkuil, et~al.
\newblock Language models enable zero-shot prediction of the effects of mutations on protein function.
\newblock {\em Advances in Neural Information Processing Systems}, 34:29287--29303, 2021.

\bibitem{mirdita2022colabfold}
M.~Mirdita, K.~Sch{\"u}tze, Y.~Moriwaki, L.~Heo, S.~Ovchinnikov, and M.~Steinegger.
\newblock Colabfold: making protein folding accessible to all.
\newblock {\em Nature methods}, 19(6):679--682, 2022.

\bibitem{norn2021protein}
C.~Norn, B.~I. Wicky, D.~Juergens, et~al.
\newblock Protein sequence design by conformational landscape optimization.
\newblock {\em Proceedings of the National Academy of Sciences}, 118(11):e2017228118, 2021.

\bibitem{pacesa2024bindcraft}
M.~Pacesa, L.~Nickel, C.~Schellhaas, J.~Schmidt, E.~Pyatova, L.~Kissling, P.~Barendse, J.~Choudhury, S.~Kapoor, A.~Alcaraz-Serna, et~al.
\newblock Bindcraft: one-shot design of functional protein binders.
\newblock {\em bioRxiv}, pages 2024--09, 2024.

\bibitem{rao2021msa}
R.~M. Rao, J.~Liu, R.~Verkuil, et~al.
\newblock Msa transformer.
\newblock {\em Proceedings of the International Conference on Machine Learning}, pages 8844--8856, 2021.

\bibitem{rives2021biological}
A.~Rives, J.~Meier, T.~Sercu, S.~Goyal, Z.~Lin, J.~Liu, D.~Guo, M.~Ott, C.~L. Zitnick, J.~Ma, et~al.
\newblock Biological structure and function emerge from scaling unsupervised learning to 250 million protein sequences.
\newblock {\em Proceedings of the National Academy of Sciences}, 118(15):e2016239118, 2021.

\bibitem{simons1997assembly}
K.~T. Simons, C.~Kooperberg, E.~Huang, and D.~Baker.
\newblock Assembly of protein tertiary structures from fragments with similar local sequences using simulated annealing and bayesian scoring functions.
\newblock {\em Journal of Molecular Biology}, 268(1):209--225, 1997.

\bibitem{bytedance2025protenix}
B.~A.~A. Team, X.~Chen, Y.~Zhang, C.~Lu, W.~Ma, J.~Guan, C.~Gong, J.~Yang, H.~Zhang, K.~Zhang, et~al.
\newblock Protenix-advancing structure prediction through a comprehensive alphafold3 reproduction.
\newblock {\em BioRxiv}, pages 2025--01, 2025.

\bibitem{voigt2000trading}
C.~A. Voigt, D.~B. Gordon, and S.~L. Mayo.
\newblock Trading accuracy for speed: A quantitative comparison of search algorithms in protein sequence design.
\newblock {\em Journal of Molecular Biology}, 299(3):789--803, 2000.

\bibitem{wang2025simplefoldfoldingproteinssimpler}
Y.~Wang, J.~Lu, N.~Jaitly, J.~Susskind, and M.~A. Bautista.
\newblock Simplefold: Folding proteins is simpler than you think, 2025.

\bibitem{watson2023novo}
J.~L. Watson, D.~Juergens, N.~R. Bennett, B.~L. Trippe, J.~Yim, H.~E. Eisenach, W.~Ahern, A.~J. Borst, R.~J. Ragotte, L.~F. Milles, et~al.
\newblock De novo design of protein structure and function with rfdiffusion.
\newblock {\em Nature}, 620(7976):1089--1100, 2023.

\end{thebibliography}

\clearpage
\appendix


\section{Experiment Details}
\subsection{Evaluation Protocol}

We follow established protein design evaluation protocols \cite{geffner2025laproteina} to assess both the quality and diversity of generated structures. Our evaluation pipeline consists of two main components: designability assessment and structural diversity analysis.

\paragraph{Designability Assessment.}
Designability measures whether a generated protein structure can be realized by at least one amino acid sequence that folds into the target conformation. This metric is particularly relevant for backbone design methods that primarily optimize structural objectives. Our evaluation proceeds as follows: (1) For each generated structure, we employ ProteinMPNN to sample $M$ candidate sequences using a temperature of 0.1, ensuring high-confidence sequence predictions. (2) Each candidate sequence is then folded using Protenix to predict its 3D structure. (3) We compute the $\alpha$-carbon RMSD between the original designed structure and each Protenix prediction. (4) A structure is classified as designable if the minimum RMSD across all $M$ sequences falls below 3.5 Å, indicating successful sequence-to-structure recovery.

We evaluate designability through two protocol variants: MPNN-1 employs a single sequence ($M=1$) for stringent evaluation, whereas MPNN-8 utilizes eight sequences ($M=8$) to capture broader designability patterns with generally improved success rates. The 3.5 Å RMSD cutoff aligns with established protein design benchmarks and reflects meaningful structural similarity for protein complexes at the backbone level.

\paragraph{Structural Diversity Analysis.}
To quantify the diversity of generated designs, we focus exclusively on structural diversity, which measures how many distinct structural modes are discovered by each method. This metric is crucial for assessing whether a design method can explore different regions of the protein fold landscape rather than generating minor variations of the same structure.

Importantly, we evaluate structural diversity only among designable samples—structures that have already passed the designability assessment described above. This ensures that diversity measurements reflect meaningful structural variation among viable protein designs rather than including potentially non-functional structures. Our analysis employs Foldseek clustering to group designable structures based on their 3D similarity, using the following command:
\begin{verbatim}
foldseek easy-cluster <samples_dir> <results_dir> <tmp_dir>
--cov-mode 0 --alignment-type 1 --min-seq-id 0 --tmscore-threshold 0.5
\end{verbatim}

This configuration clusters structures based purely on their geometric similarity (ignoring sequence information) with a TM-score threshold of 0.5, which corresponds to structures sharing similar overall folds. The number of resulting clusters serves as our structural diversity metric—higher cluster counts indicate that a method discovers more distinct structural modes within the same computational budget.

This evaluation framework enables fair comparison between RSS and baseline methods by focusing on two complementary aspects: the ability to generate structures that can be realized by natural sequences (designability) and the ability to discover diverse structural solutions (structural diversity).
\subsection{Soft-PLM}
Let \(x=(x_1,\ldots,x_L)\) be a sequence over the 20-amino-acid alphabet \(\mathcal{A}\).
For site \(i\), the discrete PLM returns a masked conditional \(p^{\mathrm{ESM}}_i(\cdot\mid x_{\backslash i})\in\Delta^{19}\) when \(x_i\) is replaced by a mask token.
Soft-PLM takes per-site probabilities \(q=(q_1,\ldots,q_L)\) with \(q_i\in\Delta^{19}\), forms expected embeddings, and outputs logits \(z^{\mathrm{soft}}_i(q)\in\mathbb{R}^{20}\) and \(p^{\mathrm{soft}}_i(\cdot\mid q)=\mathrm{softmax}\!\big(z^{\mathrm{soft}}_i(q)/\tau^\star\big)\).
A single global temperature \(\tau^\star\) is calibrated on a small split by
\[
\tau^\star \in \arg\min_{\tau>0}\ \frac{1}{M}\sum_{m=1}^{M}
D_{\mathrm{KL}}\!\left(p^{\mathrm{ESM}}_{i_m}(\cdot\mid x^{(m)}_{\backslash i_m})\ \Vert\ 
\mathrm{softmax}\!\big(z^{\mathrm{soft}}_{i_m}(q^{(m)})/\tau\big)\right),
\]
with \(q^{(m)}_j=\mathrm{onehot}(x^{(m)}_j)\) for all \(j\neq i_m\).

\paragraph{One-hot fidelity.}
For random sites \(i\), set \(q_j=\mathrm{onehot}(x_j)\) for \(j\neq i\) and evaluate
\[
D_{\mathrm{KL}}\!\left(p^{\mathrm{ESM}}_i(\cdot\mid x_{\backslash i})\ \Vert\ p^{\mathrm{soft}}_i(\cdot\mid q)\right).
\]
To check local sensitivity, define \(\mathcal{L}_i(q)= - q_i^{\top}\log p^{\mathrm{soft}}_i(\cdot\mid q)\) and, for native \(a=x_i\) and \(b\neq a\),
\[
\Delta_{a\rightarrow b} = -\log p^{\mathrm{ESM}}_i(b\mid x_{\backslash i}) + \log p^{\mathrm{ESM}}_i(a\mid x_{\backslash i}),\qquad
\delta_{a\rightarrow b} = \nabla_{q_i[b]}\mathcal{L}_i - \nabla_{q_i[a]}\mathcal{L}_i.
\]
Report the Spearman correlation between \(\{\Delta_{a\rightarrow b}\}\) and \(\{\delta_{a\rightarrow b}\}\) per site.

\paragraph{Mixture consistency.}
Create relaxed contexts by blurring a random \(30\%\) of positions:
\[
q_i^{(\varepsilon)}=(1-\varepsilon)\,\mathrm{onehot}(x_i)+\varepsilon\,\mathrm{Unif}_{20},\qquad \varepsilon\in\{0.0,0.2,0.4,0.6,0.8\}.
\]
Soft-PLM gives \(p^{\mathrm{soft}}_i(\cdot\mid q^{(\varepsilon)})\).
As a reference, draw \(K\) sequences \(x^{(k)}\sim \prod_j q^{(\varepsilon)}_j\) and average discrete PLM conditionals
\[
p^{\mathrm{MC}}_i(\cdot)=\frac{1}{K}\sum_{k=1}^{K} p^{\mathrm{ESM}}_i(\cdot\mid x^{(k)}_{\backslash i}).
\]
We measure \(\mathrm{JS}\!\left(p^{\mathrm{soft}}_i, p^{\mathrm{MC}}_i\right)\) and top-1 agreement; the implementation uses \(K=8\).

\paragraph{Library ranking.}
Pick a small site set \(S\) and option lists \(A_i\subset\mathcal{A}\) with \(|A_i|=3\).
Define the design prior \(q_i^{\mathrm{lib}}=\mathrm{Unif}(A_i)\) for \(i\in S\) and \(q_i=\mathrm{onehot}(x_i)\) elsewhere.
Soft-PLM assigns
\[
F_{\mathrm{soft}}(S,A) = \sum_{i\in S} H\!\left(q_i^{\mathrm{lib}},\,p^{\mathrm{soft}}_i(\cdot\mid q)\right).
\]
For the discrete PLM baseline, sample \(K\) full variants by choosing residues from \(A_i\) at each \(i\in S\), compute per-variant pseudo-NLL at edited sites with the discrete PLM masked conditionals, and aggregate by the mean and the best over \(K\) (\(K=256\) in our runs). Report Spearman correlations between \(F_{\mathrm{soft}}\) and each aggregate across libraries.

\paragraph{Implementation notes.}
All results use ESM2\_t30\_150M, \(N=500\) proteins, exact one-hot contexts for fidelity, temperature calibration on a small split, and fixed random seeds. The mixture and library Monte-Carlo baselines use different sample sizes for efficiency, matching the code.

\subsection{Correctness of Algorithm 1}

\begin{proposition}
The walk kernel $\mathcal{T}_{\mathrm{walk}}$ (MALA with acceptance $\alpha_{\mathrm{walk}}$ in \eqref{eq:walk-acc}) and the jump kernel $\mathcal{T}_{\mathrm{jump}}$ (PLM-guided masked swap with acceptance $\alpha_{\mathrm{jump}}$ in \eqref{eq:jump-acc}) are $\pi$-reversible. Consequently, their mixture
\[
\mathcal{T}\;=\;(1-p_{\mathrm{jump}})\,\mathcal{T}_{\mathrm{walk}}+p_{\mathrm{jump}}\,\mathcal{T}_{\mathrm{jump}}
\]
is also $\pi$-reversible and admits $\pi$ as a stationary distribution.
\end{proposition}

\textit{Proof sketch.}
For $\mathcal{T}_{\mathrm{walk}}$, Metropolis-adjusted Langevin dynamics satisfies detailed balance with target $\pi(\ell)\propto \exp(-\beta \mathcal{E}(\ell))$ by construction: the Langevin proposal density is corrected by the Metropolis–Hastings acceptance $\alpha_{\mathrm{walk}}$ in \eqref{eq:walk-acc}.

For $\mathcal{T}_{\mathrm{jump}}$, augment the state with the auxiliary variables $(S, y^{+}, y^{-})$ used to form the proposal. The deterministic update that maps $\ell$ to $\ell'$ by applying the swaps on $S$ is a bijection with unit Jacobian; its inverse swaps $y^{+}$ and $y^{-}$. The proposal density factorizes into the mask-selection term $q_S(S\mid \ell)$ and, for each $i\in S$, token draws from $p^{\mathrm{PLM}}_i(\cdot\mid \ell;\tau)$ and the base $r_i(\cdot)$ (uniform over the vocabulary). The forward/backward proposal ratio thus reduces to the product that appears in \eqref{eq:jump-acc}, and the acceptance $\alpha_{\mathrm{jump}}$ enforces detailed balance with $\pi$.

A convex combination of $\pi$-reversible kernels is $\pi$-reversible. Hence the mixture kernel $\mathcal{T}$ is $\pi$-reversible and has $\pi$ as a stationary distribution.

\section{Additional Results}
\begin{table}[t]
\centering
\caption{\textbf{Soft-PLM accurately approximates discrete PLM operations.} Comparison between Soft-PLM and ESM2\_t30\_150M on 500 UniProt proteins (average length 244 residues). Higher values are better except for KL and JS divergence.}
\begin{tabular}{l l r}
\hline
Task & Metric & Value \\
\hline
One-hot fidelity & KL divergence $D_{\mathrm{KL}}(p_{\mathrm{ESM}}\Vert p_{\mathrm{soft}})$ & 0.034 \\
                 & Gradient–swap Spearman correlation & 0.987 \\
Mixture consistency & Jensen–Shannon divergence & 0.017 \\
                    & Top-1 agreement & 0.839 \\
Library ranking & Spearman correlation (mean score) & 1.000 \\
                & Spearman correlation (best-of-$K$) & 0.842 \\
\hline
\end{tabular}
\label{tab:softesm-main}
\end{table}

\end{document}